\documentclass[sn-mathphys,iicol]{sn-jnl}

\jyear{2022}%

\usepackage{subcaption}
\usepackage{bbm}
\usepackage{enumitem}
\usepackage{bm}
\usepackage{siunitx}
\usepackage{float}

\DeclareMathOperator*{\argmin}{arg\,min}

\begin{document}

\title[Omni. Detection]{Segmentation-Based Bounding Box Generation for Omnidirectional Pedestrian Detection}


\author*[1]{\fnm{Masato} \sur{Tamura}}\email{masato.tamura.sf@hitachi.com}

\author[1]{\fnm{Tomoaki} \sur{Yoshinaga}}\email{tomoaki.yoshinaga.xc@hitachi.com}

\affil[1]{\orgname{Hitachi, Ltd.}, \orgaddress{\street{1-280, Higashikoigakubo}, \city{Kokubunji}, \postcode{185-8601}, \state{Tokyo}, \country{Japan}}}

\abstract{
We propose a segmentation-based bounding box generation method for omnidirectional pedestrian detection that enables detectors to tightly fit bounding boxes to pedestrians without omnidirectional images for training. Due to the wide angle of view, omnidirectional cameras are more cost-effective than standard cameras and hence suitable for large-scale monitoring. The problem of using omnidirectional cameras for pedestrian detection is that the performance of standard pedestrian detectors is likely to be substantially degraded because pedestrians' appearance in omnidirectional images may be rotated to any angle. Existing methods mitigate this issue by transforming images during inference. However, the transformation substantially degrades the detection accuracy and speed. A recently proposed method obviates the transformation by training detectors with omnidirectional images, which instead incurs huge annotation costs. To obviate both the transformation and annotation works, we leverage an existing large-scale object detection dataset. We train a detector with rotated images and tightly fitted bounding box annotations generated from the segmentation annotations in the dataset, resulting in detecting pedestrians in omnidirectional images with tightly fitted bounding boxes. We also develop pseudo-fisheye distortion augmentation, which further enhances the performance. Extensive analysis shows that our detector successfully fits bounding boxes to pedestrians and demonstrates substantial performance improvement.
}


\maketitle

\section{Introduction}

\begin{figure*}[tb]
	\begin{minipage}[b]{0.32\linewidth}
	\centering
	\includegraphics[keepaspectratio,scale=.77]{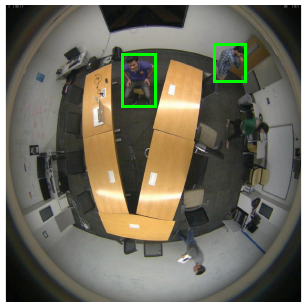}
	\end{minipage}
	\begin{minipage}[b]{0.32\linewidth}
	\centering
	\includegraphics[keepaspectratio,scale=.77]{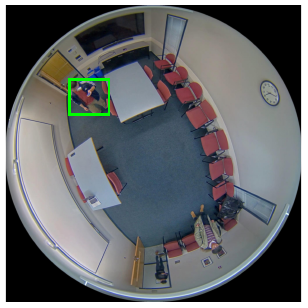}
	\end{minipage}
	\begin{minipage}[b]{0.32\linewidth}
	\centering
	\includegraphics[keepaspectratio,scale=.77]{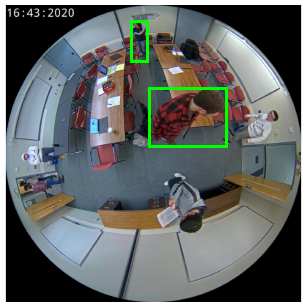}	
	\end{minipage}
	\caption{Omnidirectional pedestrian detection results of a detector trained with perspective images. The detector tends to detect only pedestrians whose appearance is nearly upright-oriented.}\label{fig:intro}
\end{figure*}
   
Pedestrian detection has spurred enormous interest owing to its diverse applications such as autonomous driving and surveillance. Following other computer vision tasks, previous works in pedestrian detection~\cite{hosang_cvpr2015,zhang_eccv2016,zhang_cvpr2017,zhang_cvpr2018,noh_cvpr2018,liu_cvpr2019,wu_cvpr2020,huan_cvpr2020,hasan_cvpr2021} leverage deep convolutional neural networks (CNNs), which demonstrate an impressive substantial improvement in performance.

As a sub-problem of pedestrian detection, omnidirectional pedestrian detection has also been studied for years~\cite{meinei_ist2014,chiang_icmew2014,cinaroglu_siu2014,krams_avss2017,demirkus_visigrapp2017,nguyen_jkms2016,seidel_visapp2019,tamura_wacv2019,li_avss2019,duan_cvprw2020,chiang_ivc2021}. A major difference between omnidirectional pedestrian detection and the general one is that pedestrians' appearance tends to be deformed due to the wide angle of view by omnidirectional cameras. This deformation is so severe in some parts of an image that it renders typical pedestrian detectors ineffective. Figure~\ref{fig:intro} shows the detection results of applying a state-of-the-art detector to omnidirectional images. Because the detector is trained with perspective images, it tends to detect only pedestrians whose appearance is nearly upright-oriented. As the wide angle of view enables cost-effective camera systems with high coverage of target areas, tremendous efforts have been dedicated to overcoming the problem and achieving high performance.

Current state-of-the-art approaches are broadly divided into two types: those that require training with omnidirectional images, and those that do not. While both approaches have achieved substantial detection accuracy, they have several drawbacks. The training-free approach~\cite{seidel_visapp2019,li_avss2019,chiang_ivc2021} transforms an omnidirectional image into multiple images and processes them with an off-the-shelf object detector. The advantage of this approach is that it does not incur annotation costs. However, because an omnidirectional image is transformed into multiple images with overlapped regions, the detection tends to be slow and generates over-detection. The approach entailing omnidirectional images for training obviates the transformation by training an object detector with omnidirectional pedestrian detection datasets. Because pedestrians' appearance in omnidirectional images is arbitrary-oriented, Duan \textit{et al.}~\cite{duan_cvprw2020} used bounding boxes with angles instead of typical upright-oriented boxes, which shows the effectiveness of the angle-aware boxes. However, even though these boxes are suitable for detection, the annotation works for rotated boxes are laborious due to the lack of the criteria for determining a box size, position, and angle, which yields multiple plausible box candidates for a pedestrian.
 
To address the aforementioned problems of the existing methods, we propose a training method that fully leverages an existing large-scale object detection dataset. In our previous conference paper~\cite{tamura_wacv2019}, we focused on the rotation of the appearance in omnidirectional images. We applied rotation data augmentation to images in the COCO object detection dataset~\cite{lin_eccv2014} during training to deal with the rotation without transformation at inference time. The diversity of the appearance, which is expanded by the rotation data augmentation, enables our detector to demonstrate remarkable performance without either the loss of inference speed or cumbersome annotations. In this paper, to further enhance the performance of omnidirectional pedestrian detection with perspective training images, we propose utilizing rotated bounding boxes generated from the segmentation annotations instead of the provided upright-oriented boxes. The generated boxes tightly fit pedestrians' appearance and hence are suitable for training omnidirectional pedestrian detectors. In addition to the bounding box generation, we propose a new data augmentation method that works well with the segmentation-based boxes. The method adds pseudo-fisheye distortion to perspective images to imitate omnidirectional ones. This imitation enhances the robustness of the detector against the distortion of omnidirectional images, which further boosts the detection performance. 

We summarize the contributions of this paper as follows:
\begin{itemize}
\item We propose generating tightly fitted bounding boxes from existing segmentation annotations for omnidirectional pedestrian detection.
\item We propose a novel data augmentation method, which enables perspective images to imitate omnidirectional ones.
\item We conduct extensive experiments on publicly available benchmark datasets and demonstrate a substantial performance improvement with the proposed method.
\end{itemize}

\section{Related work}
\subsection{Object detection}
CNNs have become a de-facto standard for object detection due to their learning capabilities and large-scale datasets. Existing CNN-based detectors are categorized into single-stage and two-stage detectors. Single-stage detectors such as YOLO~\cite{redmon_cvpr2016,redmon_cvpr2017,redmon_arxiv2018,bochkovskiy_arxiv2020,wang_cvpr2021,ge_arxiv2021} achieve high-speed detection with competitive detection accuracy, while two-stage ones such as Faster R-CNN~\cite{ren_nips2015} acquire high accuracy at the cost of large computations.

Considering the attractive performance and straightforward implementation, most existing omnidirectional pedestrian detection methods~\cite{nguyen_jkms2016,seidel_visapp2019,li_avss2019,duan_cvprw2020,chiang_ivc2021}, as well as ours in the conference paper~\cite{tamura_wacv2019}, adopt YOLO as a base detector. YOLO uses pixel-wise dense prediction, which outputs duplicated detection results, and thus requires a suppression post-process. Non-maximum suppression (NMS) is typically used for this suppression, which calculates intersection over union (IoU) to determine whether boxes should be suppressed. However, because the calculation of IoU for a pair of rotated bounding boxes is quite complicated, NMS is inappropriate for omnidirectional pedestrian detection.

A detector named DETR~\cite{carion_eccv2020}, which does not need the suppression, and its extended version Deformable DETR~\cite{zhu_iclr2020} have been proposed. The authors of DETR formulate object detection as a set prediction and train a detector with the bipartite matching between predictions and ground truths. Leveraging Transformer's strong feature extraction capability~\cite{vaswani_nips2017}, DETR has achieved competitive performance to Faster R-CNN without a cumbersome post-process. The authors of Deformable DETR tackle the problems of slow convergence and limited feature spatial resolution in DETR and have achieved superior performance in a shorter training time. In this paper, to take advantage of post-process-free and high-performance detection, we use Deformable DETR as a base detector and extend it for omnidirectional pedestrian detection.

\subsection{Omnidirectional pedestrian detection}
Due to the lack of large-scale omnidirectional pedestrian detection datasets, early attempts relied on handcrafted features~\cite{meinei_ist2014,chiang_icmew2014,cinaroglu_siu2014,demirkus_visigrapp2017,krams_avss2017}. The handcrafted feature-based methods approximate the features of ordinary pedestrians' appearance from the deformed appearance by transforming their images or features. This approximation enables classifiers trained with perspective images to be utilized. However, because the deformation of the appearance is prominent in some parts of an image, approximation errors are inevitable, causing performance degradation.

Another approach to solving the lack of large-scale datasets is to utilize CNN-based object detectors trained with perspective images. Seidel \textit{et al.}~\cite{seidel_visapp2019} proposed transforming an omnidirectional image into multiple perspective images by projection transformation with calibrated camera parameters. YOLOv2~\cite{redmon_cvpr2017}, which is trained with the COCO dataset~\cite{lin_eccv2014}, is applied to the transformed images, and then the detection results are projected back to the omnidirectional image. Instead of the projection transformation, Li \textit{et al.}~\cite{li_avss2019} proposed rotating an omnidirectional image at an interval and applying YOLOv3~\cite{redmon_arxiv2018}, which is also trained with the COCO dataset, to a window in the rotated image. The detection results are projected back to their original positions. Because these methods utilize the detectors trained with the existing dataset, they mitigate the cumbersome data preparation for training detectors. However, they tend to show over-detection because of the overlapped boundary regions and slow detection speed due to processing multiple images transformed from an omnidirectional image. Chiang \textit{et al.}~\cite{chiang_ivc2021} proposed dividing an omnidirectional image into multiple patches, transforming them, and combining them into one image. This method mitigates the requirement of processing multiple images generated from an omnidirectional image during inference. However, the method still has the problem of over-detection and requires transformation pre-processing and projection post-processing.

To solve the problems of the transformation during inference, Duan \textit{et al.}~\cite{duan_cvprw2020} proposed training a detector with several omnidirectional pedestrian detection datasets annotated with rotated bounding boxes. An angle-aware loss is introduced into YOLOv3 to predict the angles of boxes and tightly fit them to pedestrians. Owing to the angle-aware prediction and datasets, a substantial performance has been demonstrated. However, the annotations for this method are laborious because the ground truths of rotated boxes are not uniquely determined.

To leverage their angle-aware detection without laborious annotation works, we extend our previously proposed method~\cite{tamura_wacv2019}. In the previous method, images and upright-oriented bounding box annotations in the COCO dataset are rotated during training to make YOLOv2 rotationally invariant. The trained YOLOv2 outputs actual heights and widths of pedestrians' appearance in omnidirectional images regardless of their angles. Their positions determine the angles of boxes under the assumption that pedestrians' appearance in omnidirectional images is typically in radial directions. This method makes it possible to detect pedestrians without either the transformation at inference time or using omnidirectional images for training. However, since the detector is trained with upright-oriented boxes and does not predict angles, predicted boxes do not typically fit pedestrians. To solve this problem, we utilize segmentation-based bounding box annotations with angle-aware detection.

\section{Proposed method}
\subsection{Overview of the detector}
\begin{figure}[tb]
 \begin{center}
 \centerline{\includegraphics[width=1.\linewidth]{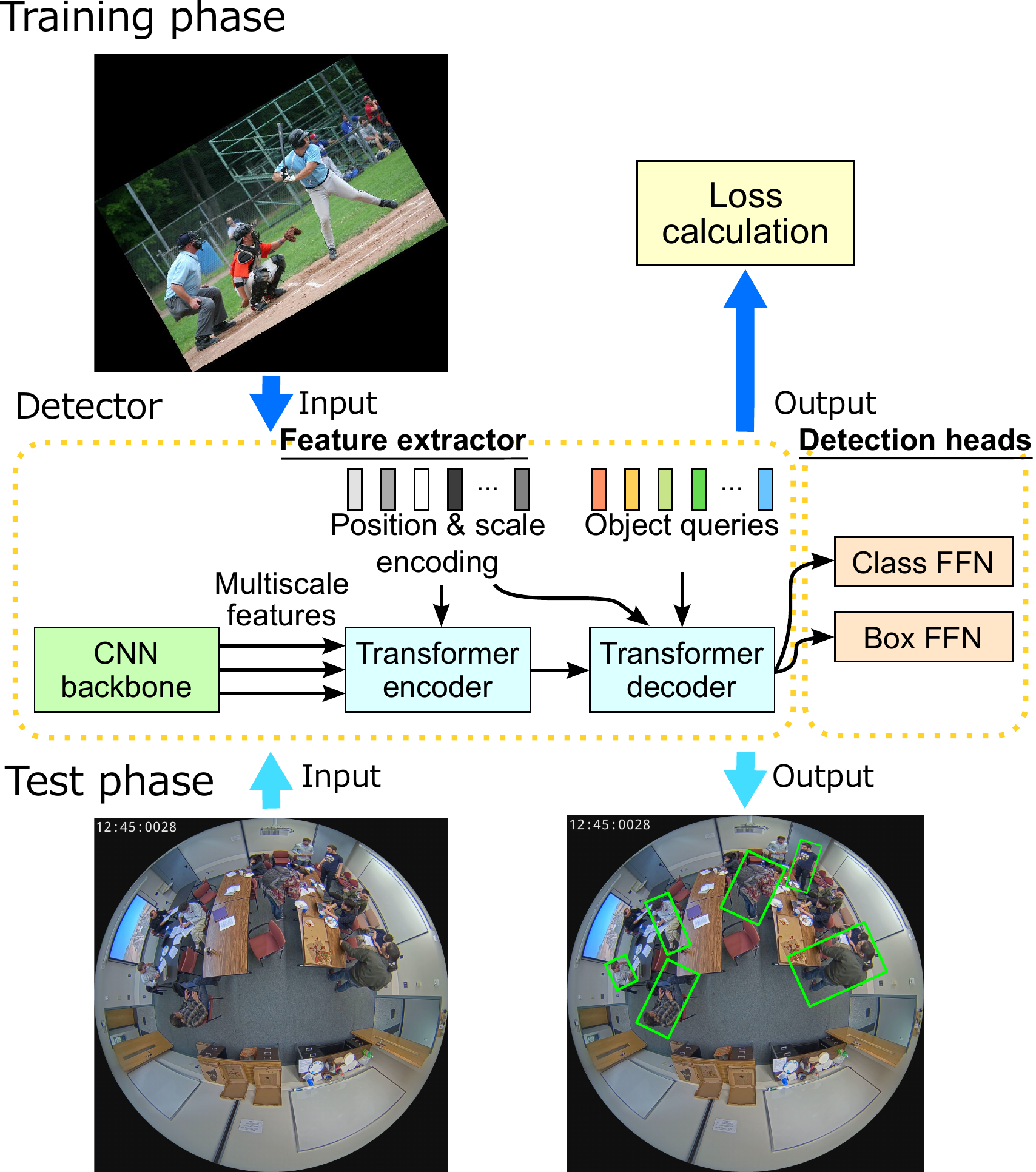}}
 \caption{The overall architecture of the proposed method. During training, perspective images are input to the detector network, while during evaluation, omnidirectional images are input.}
 \label{fig:arch}
 \end{center}
\end{figure} 

To leverage the rotation-aware~\cite{duan_cvprw2020} and suppression-free~\cite{carion_eccv2020,zhu_iclr2020} detection for omnidirectional pedestrian detection, we train Deformable DETR with the angle-aware loss function. We briefly review the overall architecture and loss calculation for training the detector.

Figure~\ref{fig:arch} shows the architecture of our omnidirectional pedestrian detector. Given an input image $\bm{x} \in \mathbb{R}^{3 \times H \times W}$, multi-scale feature maps $\bm{Z} = \left\{\bm{z}_i \mid \bm{z}_{i} \in \mathbb{R}^{D_z \times H_{i}^{\prime} \times W_{i}^{\prime}}\right\}_{i=1}^{L}$ are extracted by an arbitrary CNN backbone (e.g., ResNet~\cite{he_cvpr2016}) with extra projection convolution layers, where $H$ and $W$ are the height and width of the input image, respectively, $H_{i}^{\prime}$ and $W_{i}^{\prime}$ are those of each output feature map, respectively, $D_z$ is the number of projected feature map channels, and $L$ is the number of scales.

The transformer takes the feature maps $\bm{Z}$ and transforms a set of learnable object query vectors $\bm{Q} = \left\{\bm{q}_{i} \mid \bm{q}_{i} \in \mathbb{R}^{D_z} \right\}_{i=1}^{N_q}$ into a set of object embeddings $\bm{D} = \left\{\bm{d}_{i} \mid \bm{d}_{i} \in \mathbb{R}^{D_z} \right\}_{i=1}^{N_q}$ with the multi-head attention mechanism~\cite{vaswani_nips2017}, where $N_q$ is the number of queries.

The subsequent detection heads further process the embeddings to produce prediction results. The class feed-forward network (FFN), whose outputs are processed with the sigmoid function, predicts a set of the probabilities of object classes $\bm{\hat{C}} = \left\{\bm{\hat{c}}_i \mid \bm{\hat{c}}_i \in [0, 1]^{N_{obj}}\right\}_{i=1}^{N_q}$, where $N_{obj}$ is the number of the object classes. For omnidirectional pedestrian detection, $N_{obj}$ is set to $1$. The box FFN, whose outputs are processed with the sigmoid function, predicts normalized bounding boxes with normalized angles $\bm{\hat{B}} = \left\{\left(\bm{\hat{b}}_i, \hat{a}_i \right) \mid \bm{\hat{b}}_i \in [0, 1]^{4}, \hat{a}_i \in [0, 1] \right\}_{i=1}^{N_q}$.

To train the detector with the bipartite matching between predictions and ground truths, we follow the procedure of DETR~\cite{carion_eccv2020} and use the Hungarian algorithm~\cite{kuhn_1955} to match the predictions and ground truths. The ground truths are padded with $\phi$ (no pedestrians) so that their sizes become $N_q$. After the Hungarian matching, we acquire the optimal assignment $\hat{\omega}$ among the set of all possible permutations of $N_q$ elements $\bm{\Omega_{N_q}}$, i.e., $\hat{\omega} = \argmin_{\omega \in \bm{\Omega}_{N_q}}{\sum_{i=1}^{N_q}{\mathcal{H}_{i,\omega(i)}}}$, where $\mathcal{H}_{i,j}$ is the matching cost between the $i$-th ground truth and $j$-th prediction. This cost is calculated in the same way as the loss calculation for each prediction and ground truth pair described as follows.

Suppose we have padded ground-truth one-hot object-class labels $\bm{C} = \left\{\bm{c}_i \mid \bm{c}_i \in \left\{0, 1\right\}^{N_{obj}}\right\}_{i=1}^{N_q}$ and normalized bounding boxes with angles $\bm{B} = \left\{\left(\bm{b}_i, \theta_i \right) \mid \bm{b}_i \in [0, 1]^{4}, \theta_i \in [-\frac{\pi}{2}, \frac{\pi}{2}) \right\}_{i=1}^{N_q}$. Following Duan \textit{et al.} \cite{duan_cvprw2020}, we restrict the angles of the boxes to be between $-\frac{\pi}{2}$ and $\frac{\pi}{2}$, and the heights of the boxes to be larger than the widths. The loss for all the matched pairs of the predictions and ground truths $\mathcal{L}$ is calculated as follows.

\begin{align}
 \mathcal{L} ={} & \lambda_{c} \mathcal{L}_{c} + \lambda_{b} \mathcal{L}_{b} + \lambda_{u} \mathcal{L}_{u} + \lambda_{a} \mathcal{L}_{a}, \\
 \nonumber\mathcal{L}_{c} ={} & \frac{1}{\lvert\bar{\bm{\Phi}}\rvert} \sum_{i=1}^{N_{q}}\left\{
 \mathbbm{1}_{\{i \not\in \bm{\Phi}\}}\left[l_{f}\left(\bm{c}_i, \bm{\hat{c}_{\hat{\omega}\left(i\right)}}\right)\right]\right.\\
 &{} \qquad\qquad\qquad \left. + \mathbbm{1}_{\{i \in \bm{\Phi}\}}\left[l_{f}\left(\bm{0}, \bm{\hat{c}_{\hat{\omega}\left(i\right)}}\right)\right]
 \right\}, \\
 \mathcal{L}_{b} ={} & \frac{1}{\lvert\bar{\bm{\Phi}}\rvert} \sum_{i=1}^{N_{q}} \mathbbm{1}_{\{i \not\in \bm{\Phi}\}} \left[
 \left\|\bm{b}_i - \bm{\hat{b}}_{\hat{\omega}\left(i\right)}\right\|_{1}\right], \\
 \mathcal{L}_{u} ={} & \frac{1}{\lvert\bar{\bm{\Phi}}\rvert} \sum_{i=1}^{N_{q}} \mathbbm{1}_{\{i \not\in \bm{\Phi}\}}\left[1 - 
 \mathrm{GIoU}\left(\bm{b}_i, \bm{\hat{b}}_{\hat{\omega}\left(i\right)}\right)\right], \\
 \mathcal{L}_{a} ={} & \frac{1}{\lvert\bar{\bm{\Phi}}\rvert} \sum_{i=1}^{N_{q}} \mathbbm{1}_{\{i \not\in \bm{\Phi}\}} \\
 &{} \qquad \left[\left\|\bmod\left(\hat{\theta}_{\hat{\omega}\left(i\right)} - \theta_i -\frac{\pi}{2}, \pi\right) - \frac{\pi}{2} \right\|_{1}\right],
\end{align}
where $\hat{\theta}_i = 2\pi\hat{a}_i - \pi$, $\bm{\Phi}$ is a set of ground-truth indices that correspond to $\phi$, $\mathrm{GIoU}(\cdot, \cdot)$ is the generalized IoU function~\cite{rezatofighi_cvpr2019}, $l_f\left(\cdot, \cdot\right)$ is the focal loss function~\cite{lin_iccv2017}, and $\lambda_{\{c, b, u, a\}}$ are the hyper-parameters for adjusting the weight of each loss. Because of these loss designs, the detector learns to assign one query to at most one object in an image without duplication, and as a result, the suppression of prediction results can be obviated.

\subsection{Bounding box generation from segmentation annotations}
\begin{figure}[tb]
 \begin{minipage}[b]{1.0\linewidth}
 \centering
 \includegraphics[keepaspectratio,width=.6\linewidth]{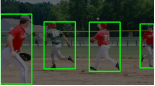}
 \subcaption{Provided bounding box.}\label{fig:anno_bbox}
 \end{minipage} \\
 \begin{minipage}[b]{1.0\linewidth}
 \centering
 \includegraphics[keepaspectratio,width=0.6\linewidth]{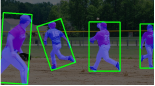}
 \subcaption{Segmentation-based bounding box.}\label{fig:anno_seg}
 \end{minipage}
 \caption{Provided and generated bounding boxes.}\label{fig:anno}
\end{figure}

Duan \textit{et al.}~\cite{duan_cvprw2020} demonstrated that angle-aware detection can tightly fit bounding boxes to pedestrians and thus is suitable for omnidirectional pedestrian detection. However, because the box annotations of the COCO dataset~\cite{lin_eccv2014} are upright-oriented and do not tightly fit pedestrians, angle-aware detection cannot fully be leveraged. To overcome this problem, we generate boxes from the segmentation annotations in the COCO dataset and use the generated boxes instead of the original ones to train the detector.

Suppose we have segmentation annotations for pedestrians $\mathcal{S} = \left\{\bm{S}_i\right\}_{i = 1}^{N_p}$, where $\bm{S}_i = \left\{\bm{p}_j^{(i)} \mid \bm{p}_j^{(i)} \in \mathbb{R}^{2} \right\}_{j = 1}^{N_i^{(s)}}$ is a segment for a pedestrian, $\bm{p}_j^{(i)}$ is a vertex of the segment, $N_p$ is a number of pedestrians, and $N_i^{(s)}$ is a number of vertices for the segment. Taking $\mathcal{S}$, rotated bounding boxes $\tilde{\bm{B}} = \left\{\left(\tilde{\bm{b}}_i, \theta_i \right) \mid \tilde{\bm{b}}_i \in \mathbb{R}^{4}, \theta_i \in [-\frac{\pi}{2}, \frac{\pi}{2}) \right\}_{i=1}^{N_p}$ are generated as follows.
\begin{equation}\label{eq:mbr}
 \tilde{\bm{B}} = f_{MBR}\left(f_{hull}\left(\mathcal{S}\right)\right),
\end{equation}
where $f_{hull}\left(\cdot\right)$ is a function that finds a convex hull for each segment, and $f_{MBR}\left(\cdot\right)$ is a function that finds a minimum bounding rectangle for each convex hull. During the training, the generated boxes are normalized with image sizes and used for loss calculation.

Examples of original upright-oriented and generated bounding boxes are illustrated in Fig.~\ref{fig:anno}. The segmentation region is highlighted with blue in Fig.~\ref{fig:anno_seg}. As shown in the figure, the generated boxes more tightly fit the pedestrians than the original ones. By utilizing the generated boxes, the detector learns to tightly fit boxes to pedestrians.

\subsection{Pseudo-fisheye distortion augmentation}
\begin{figure}[tb]
 \begin{minipage}[b]{1.0\linewidth}
 \centering
 \includegraphics[keepaspectratio,scale=1.4]{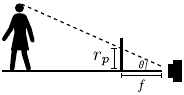}
 \subcaption{Perspective projection.}\label{fig:projection_perspective}
 \end{minipage}\\
 \begin{minipage}[b]{1.0\linewidth}
 \centering
 \includegraphics[keepaspectratio,scale=1.4]{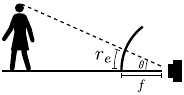}
 \subcaption{Equidistant projection.}\label{fig:projection_omni}
 \end{minipage}
 \caption{Projection methods.}\label{fig:projection}
\end{figure}

\begin{figure}[tb]
 \begin{center}
 \centerline{\includegraphics[width=0.7\linewidth]{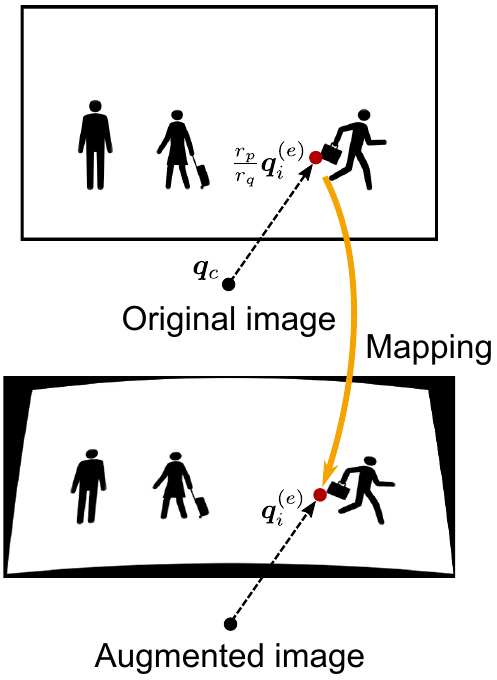}}
 \caption{Image mapping from perspective to equidistant projection.}
 \label{fig:mapping}
 \end{center}
\end{figure} 

To add pseudo-fisheye distortion to images in the COCO dataset, we consider that all the images are made by the perspective projection shown in Fig.~\ref{fig:projection_perspective} and transform them into those with the equidistant projection shown in Fig.~\ref{fig:projection_omni}. This augmentation can be applied with a combination of the segmentation-based bounding boxes because segmentation annotations can be precisely projected to transformed images, while the original boxes are distorted and do not fit pedestrians in transformed images.

In the perspective projection shown in Fig.~\ref{fig:projection_perspective}, the distance between the projection point and the optical axis on the screen $r_{p}$ is calculated as follows.
\begin{equation}\label{eq:proj_equi}
 r_p = f \tan \theta,
\end{equation}
where $f$ is the focal length and $\theta$ is the angle between the line of the projected point and optical axis. In the equidistant projection shown in Fig.~\ref{fig:projection_omni}, the distance between the projection point and the optical axis on the screen $r_{e}$ is calculated as follows.
\begin{equation}\label{eq:proj_pers}
 r_e = f \theta.
\end{equation}
Utilizing Eq.~\ref{eq:proj_equi} and~\ref{eq:proj_pers}, pixel values in a transformed image are mapped from an original image as illustrated in Fig.~\ref{fig:mapping}. We can obtain the mapping for points $\bm{Q}_e = \left\{\bm{q}_{i}^{(e)} \mid \bm{q}_{i}^{(e)} \in \mathbb{Z}^2\right\}_{i=1}^{N_e}$ on the grid of a transformed image as follows.
\begin{align}
 \nonumber\bm{q}_{i}^{(p)} =&{} \frac{r_p}{r_e}\bm{q}_{i}^{(e)} + \bm{q}_c \\
 \nonumber=&{} \frac{f\tan \left(\frac{\|\bm{q}_{i}^{(e)}\|_{2}}{f}\right)}{\|\bm{q}_{i}^{(e)}\|_{2}}\bm{q}_{i}^{(e)} + \bm{q}_c \\
 &{} \qquad \mbox{ s.t. } \bm{q}_{i}^{(p)} \in \bm{I},
\end{align}
where $\bm{q}_{i}^{(p)} \in \mathbb{R}^2$ is a source point in an original image, $\bm{q}_c$ is the position of the optical axis in the image coordinates, and $\bm{I}$ is the image region. The pixel values at point $\bm{q}_{i}^{(e)}$ is determined by sampling the values at point $\bm{q}_{i}^{(p)}$. Note that if point $\bm{q}_{i}^{(p)}$ is not on the grid of an original image, we sample the values with the bilinear interpolation. $f$ and $\bm{q}_c$ are randomly determined each time the augmentation is applied. 

For the bounding box generation in transformed images, we first map the vertices of a segment into a transformed image and then generate boxes. We determine the point $\bm{q}_{l}^{(e)}$ in a transformed image for the vertex $\bm{p}_j^{(i)}$ of a segment as follows.
\begin{equation}
 l = \argmin_{k}\left\|\bm{q}_{k}^{(p)} - \bm{p}_j^{(i)}\right\|_2.
\end{equation}
After all the vertices in a segment are mapped into a transformed image, we apply Eq.~\ref{eq:mbr} to the segment and generate a bounding box.

\section{Experiments}
\subsection{Datasets and evaluation settings}\label{subsec:dataset}
\begin{table}[t]
 \caption{Characteristics of omnidirectional pedestrian detection datasets.}
 \label{table:dataset}
 \centering
 \begin{tabular}{@{}lccc@{}}
 \toprule
 Dataset & \#Scenes & \#Frames & \#Instances \\
 \midrule
 MW-18Mar & 19 & 8,751 & 22,826 \\
 HABBOF & 4 & 5,837 & 20,467 \\
 CEPDOF & 8 & 25,504 & 173,074\\
 \bottomrule
 \end{tabular}
\end{table}

To evaluate the effectiveness of our method, we conducted experiments using the COCO object detection dataset~\cite{lin_eccv2014}, MW-18Mar\footnote{Available at \url{http://www2.icat.vt.edu/mirrorworlds/challenge/index.html}}, HABBOF~\cite{li_avss2019}, and CEPDOF~\cite{duan_cvprw2020}. The latter three datasets are omnidirectional pedestrian detection datasets. Unless otherwise stated, the COCO dataset is used for the training and the other datasets are used for the evaluation. We create a training set from the train2017 set of the COCO annotation by selecting images that have at least one person label. The created training set has 64,115 images and 262,465 person instances. The characteristics of the omnidirectional pedestrian detection datasets are described in Table~\ref{table:dataset}. For the ground truths of these three datasets, we use tightly fitted bounding boxes provided by Li \textit{et al.} \cite{li_avss2019} and Duan \textit{et al.} \cite{duan_cvprw2020}.

For the evaluation metrics, we use average precision (AP). Depending on IoU thresholds, we report three types of AP. The first is averaged over IoU thresholds from 0.5 to 0.95, which is indicated by AP. The second is with an IoU threshold of 0.5, which is indicated by $\mathrm{AP}_{50}$. The last is with an IoU threshold of 0.75, which is indicated by $\mathrm{AP}_{75}$. AP with high IoU thresholds means the performance of the detectors for fitting bounding boxes to ground truths.

\subsection{Implementation details}
We implement our method by extending Deformable DETR~\cite{zhu_iclr2020} and using ResNet-50~\cite{he_cvpr2016} as the CNN backbone network. Both the transformer encoder and decoder consist of six transformer layers with multi-head deformable attention layers of eight heads. The number of projected feature map channels $D_z$ is set to 256, and the number of query vectors $N_q$ is set to 300. The class FFN has a linear layer, while the box FFN has three linear layers with ReLU activations.

For training, we initialize the ResNet50 backbone with the parameters pre-trained on ImageNet~\cite{deng_cvpr2009}. The detector is trained for 50 epochs, and the learning rate is decayed by 0.1 after 40 epochs. Eight NVIDIA Tesla V100 GPUs are used, each of which processes a batch size of 4 at each iteration. We use the AdamW optimizer~\cite{loshchilov_iclr2019} with a base learning rate of $2 \times 10^{-4}, \beta_{1} = 0.9, \beta_{2} = 0.999$, and a weight decay of $10^{-4}$. The hyper-parameters for the loss weights $\lambda_c, \lambda_b, \lambda_u$, and $\lambda_a$ are set to 2, 5, 2, and 0.1, respectively. We set the weight for the angle loss $\lambda_a$ to a relatively lower value than the others. We analyze the effect of this weight value in Sec.~\ref{subsubsec:angle_cost}. As the default data augmentation, we add random rotation and color jitter augmentation to the augmentation of the DETR training~\cite{carion_eccv2020}.

\subsection{Quantitative comparisons}
\begin{table*}[t]
	\caption{Comparison against state-of-the-art methods. (S: The detector is trained with bounding boxes generated from segmentation annotations, D: Training images are transformed by the proposed pseudo-fisheye distortion augmentation.)}
	\label{table:comp}
 \addtolength{\tabcolsep}{-2pt}
 \centering
	\begin{tabular}[t]{@{}lccccccccc@{}}
		\toprule
		& \multicolumn{3}{c}{MW-18Mar} & \multicolumn{3}{c}{HABBOF} & \multicolumn{3}{c}{CEPDOF} \\
 \cmidrule(lr){2-4}\cmidrule(lr){5-7}\cmidrule(lr){8-10}
 & $\mathrm{AP}$ & $\mathrm{AP}_{50}$ & $\mathrm{AP}_{75}$ & $\mathrm{AP}$ & $\mathrm{AP}_{50}$ & $\mathrm{AP}_{75}$ & $\mathrm{AP}$ & $\mathrm{AP}_{50}$ & $\mathrm{AP}_{75}$ \\
 \midrule
 Seidel \textit{et al.}~\cite{seidel_visapp2019} & 23.1 & 57.3 & 10.9 & 23.3 & 59.8 & 14.3 & 18.8 & 49.4 & 8.5 \\
 Our previous~\cite{tamura_wacv2019} & 34.1 & 86.3 & 18.5 & 33.5 & 86.6 & 14.7 & 25.8 & 69.8 & 9.9 \\
 RAPiD~\cite{duan_cvprw2020} & 35.8 & 92.0 & 18.7 & 32.5 & 88.5 & 8.5 & 28.8 & 76.4 & 8.1 \\
 \midrule
 Ours (S) & 49.5 & 92.9 & \textbf{46.8} & \textbf{48.7} & 92.0 & \textbf{44.8} & 33.2 & 76.6 & 19.9 \\
 Ours (S + D) & \textbf{50.6} & \textbf{94.8} & \textbf{46.8} & 46.8 & \textbf{92.4} & 39.4 & \textbf{33.4} & \textbf{77.5} & \textbf{20.2} \\
		\bottomrule
	\end{tabular}
 \addtolength{\tabcolsep}{3pt}
\end{table*}

To show the effectiveness of the proposed method, we first compare our method with three baseline methods. For a fair comparison, we implement all methods by extending Deformable DETR~\cite{zhu_iclr2020}. As our proposed method, all the detectors in the baseline methods are trained with the COCO training set described in Sec.~\ref{subsec:dataset}. The three methods are as follows.
\begin{description}[font=\bfseries,style=unboxed,leftmargin=0cm]
 \item[Seidel \textit{et al.}~\cite{seidel_visapp2019}] We select this baseline as a representative for image transformation methods. This method first transforms an omnidirectional image into multiple perspective images with a camera calibration method and then applies a detector, which is trained with perspective images, to the transformed images. The detection results are projected back to the omnidirectional image and suppressed by soft NMS~\cite{bodla_iccv2017}. Because the dataset authors do not provide the calibration parameters, we estimate the parameters by pedestrians' appearance in the images following~\cite{tamura_wacv2019}.
 \item[Our previous work~\cite{tamura_wacv2019}] This method trains detectors with upright-oriented bounding boxes with rotation data augmentation. The trained detectors output bounding boxes with actual heights and widths of pedestrians' appearance in omnidirectional images regardless of their angles. Their positions determine the angles under the assumption that the appearance of pedestrians is typically in radial directions.
 \item[RAPiD~\cite{duan_cvprw2020}] Instead of using the position-dependent angles in our previous work, this method predicts angles of bounding boxes to precisely fit bounding boxes. Note that the original work uses omnidirectional datasets for training, while in our experiments, the detector is trained with only the COCO dataset because our focus is to achieve high performance without laborious annotation works.
\end{description}

Table~\ref{table:comp} shows the comparison results. We evaluate the performance of two variants for our method. One is that the detector is trained with the segmentation-based bounding boxes but without the proposed pseudo-fisheye distortion, which is indicated by ``S" in the table. The other is that the detector is trained with both the segmentation-based bounding boxes and pseudo-fisheye distortion, which is indicated by ``S + D". As seen from the table, the proposed bounding box generation method outperforms the baseline methods. In particular, the generation method demonstrates a substantially higher performance than the other methods when the IoU threshold is 0.75. The $\mathrm{AP}_{75}$ is improved by 28.1 on the MW-18Mar dataset, 36.3 on the HABBOF dataset, and 11.8 on the CEPDOF dataset from RAPiD. These results indicate that the proposed generation method is crucial to tightly fit boxes to pedestrians. This observation is further confirmed qualitatively in Sec.~\ref{sec:qualitative}.

The results show that the proposed distortion further improves the performance of the detector on the MW-18Mar and CEPDOF datasets. Because they have more scenes and frames than the HABBOF dataset as shown in Table~\ref{table:dataset}, the results of the former two are more reliable than those of the HABBOF dataset. Therefore, the proposed distortion is effective in training omnidirectional pedestrian detectors with perspective images.

\subsection{Qualitative comparisons}\label{sec:qualitative}
\begin{figure*}
	\centering\begin{tabular}{@{}c@{ }c@{ }c@{ }c@{ }c@{}}
	&Seidel \textit{et al.}~\cite{seidel_visapp2019} & Our previous~\cite{tamura_wacv2019} & RAPiD~\cite{duan_cvprw2020} & Ours \\
	\rotatebox{90}{\hspace{1.6em}MW-18Mar}&
	\includegraphics[width=.22\linewidth]{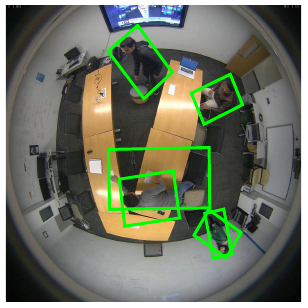}&
	\includegraphics[width=.22\linewidth]{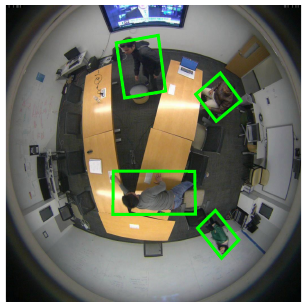}&
	\includegraphics[width=.22\linewidth]{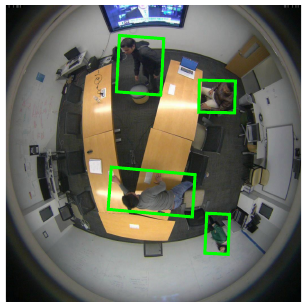}&
	\includegraphics[width=.22\linewidth]{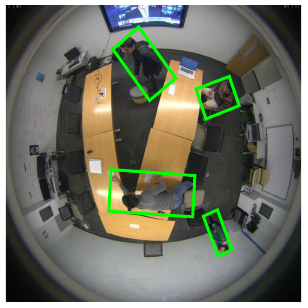}\\[-1ex]
	&(a)\label{fig:quality_mw1_pers}&(b)\label{fig:quality_mw1_rot}&(c)\label{fig:quality_mw1_rap}&(d)\label{fig:quality_mw1_our} \\
	\rotatebox{90}{\hspace{1.6em}MW-18Mar}&
	\includegraphics[width=.22\linewidth]{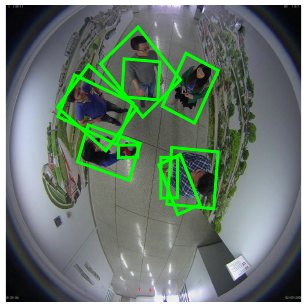}&
	\includegraphics[width=.22\linewidth]{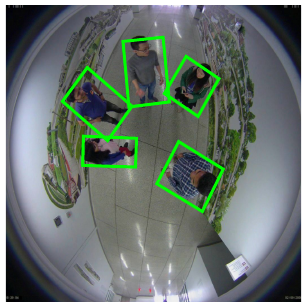}&
	\includegraphics[width=.22\linewidth]{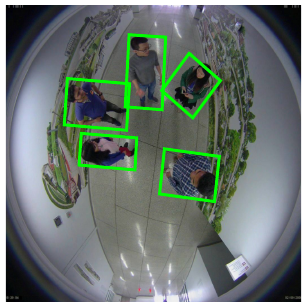}&
	\includegraphics[width=.22\linewidth]{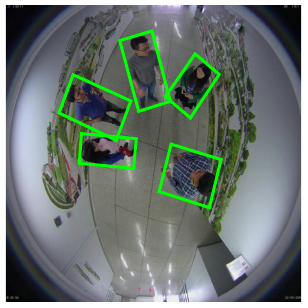}\\[-1ex]
	&(e)\label{fig:quality_mw2_pers}&(f)\label{fig:quality_mw2_rot}&(g)\label{fig:quality_mw2_rap}&(h)\label{fig:quality_mw2_our} \\
	\rotatebox{90}{\hspace{2.0em}HABBOF}&
	\includegraphics[width=.22\linewidth]{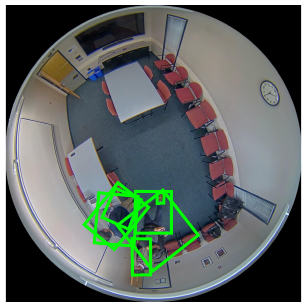}&
	\includegraphics[width=.22\linewidth]{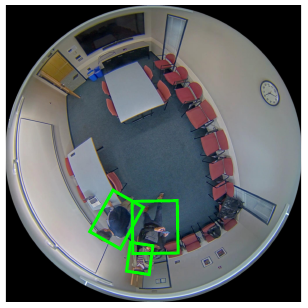}&
	\includegraphics[width=.22\linewidth]{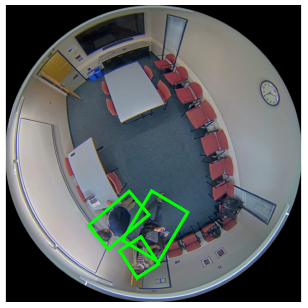}&
	\includegraphics[width=.22\linewidth]{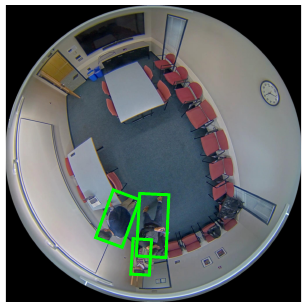}\\[-1ex]
	&(i)\label{fig:quality_habbof_pers}&(j)\label{fig:quality_habbof_rot}&(k)\label{fig:quality_habbof_rap}&(l)\label{fig:quality_habbof_our}\\
	\rotatebox{90}{\hspace{2.0em}CEPDOF}&
	\includegraphics[width=.22\linewidth]{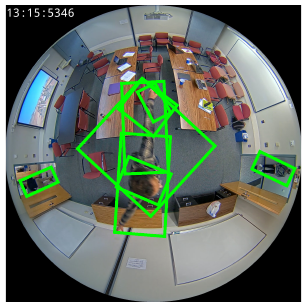}&
	\includegraphics[width=.22\linewidth]{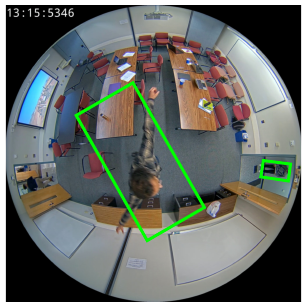}&
	\includegraphics[width=.22\linewidth]{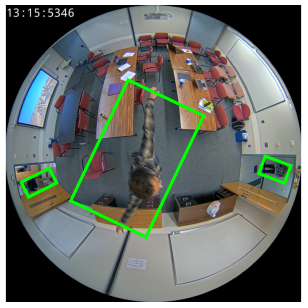}&
	\includegraphics[width=.22\linewidth]{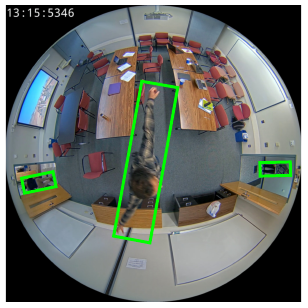}\\[-1ex]
	&(m)\label{fig:quality_cepdof1_pers}&(n)\label{fig:quality_cepdof1_rot}&(o)\label{fig:quality_cepdof1_rap}&(p)\label{fig:quality_cepdof1_our}\\
	\rotatebox{90}{\hspace{2.0em}CEPDOF}&
	\includegraphics[width=.22\linewidth]{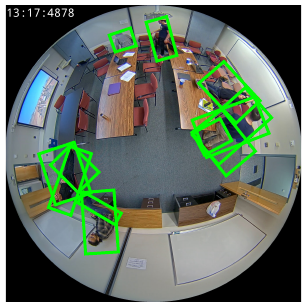}&
	\includegraphics[width=.22\linewidth]{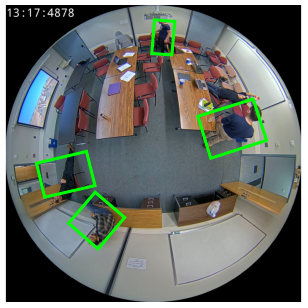}&
	\includegraphics[width=.22\linewidth]{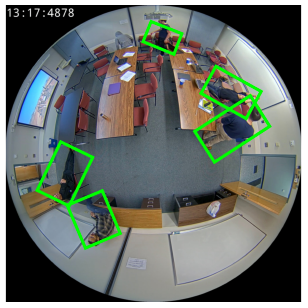}&
	\includegraphics[width=.22\linewidth]{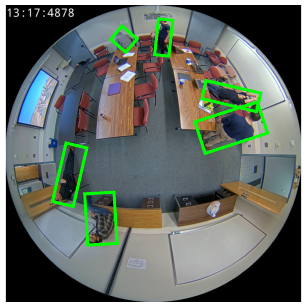}\\[-1ex]
	&(q)\label{fig:quality_cepdof2_pers}&(r)\label{fig:quality_cepdof2_rot}&(s)\label{fig:quality_cepdof2_rap}&(t)\label{fig:quality_cepdof2_our}\\
	\end{tabular}
	\caption{Example detection results of the baseline and proposed methods for each dataset. The results of our method are those of the detector trained with the segmentation-based bounding boxes and pseudo-fisheye distortion.}%
	\label{fig:quality}
\end{figure*}

To qualitatively confirm the effectiveness of the proposed method, we analyze the detection results of the baseline methods and the proposed segmentation-based method with the pseudo-fisheye distortion on the three datasets. Figure~\ref{fig:quality} illustrates the detection results. We draw bounding boxes that have scores higher than 0.5. The images in each column show the results of each method.

As shown in the figures, the proposed method fits the bounding boxes to pedestrians more tightly than the baseline methods. The method of Seidal \textit{et al.} tends to generate over-detection as shown in Fig.~\hyperref[fig:quality_mw1_pers]{5a}, \hyperref[fig:quality_mw2_pers]{5e}, \hyperref[fig:quality_habbof_pers]{5i}, \hyperref[fig:quality_cepdof1_pers]{5m}, and \hyperref[fig:quality_cepdof2_pers]{5q} because the detector need to detect in overlapped regions to cover all the areas of an image. In the results of our previous work and RAPiD, pedestrians' appearance occupies only half of the boxes in a number of cases, as can be seen with the pedestrian at the top of Fig.~\hyperref[fig:quality_mw1_rot]{5b} and \hyperref[fig:quality_mw1_rot]{5c}, and that in the middle of Fig.~\hyperref[fig:quality_cepdof1_rot]{5n} and~\hyperref[fig:quality_cepdof1_rot]{5o}. These results are caused by just rotating the original upright-oriented boxes during training, which cannot tightly fit pedestrians whose appearance is not upright. Compared with the detected boxes of the baseline methods, those of the proposed method are appropriately rotated to the angles of the appearance and fitted to the pedestrians, as can be seen with the pedestrian at the top of Fig.~\hyperref[fig:quality_mw1_our]{5d} and that in the middle of Fig.~\hyperref[fig:quality_cepdof1_our]{5p}. These results demonstrate that the proposed method can tightly fit boxes to pedestrians in omnidirectional images without using the images during training.

\subsection{Ablation experiments}
\subsubsection{Analysis of angle cost}\label{subsubsec:angle_cost}
\begin{table}
	\caption{Performance with different weights for the angle loss. The performance is evaluated using the detector trained with the segmentation-based bounding boxes and pseudo-fisheye distortion on the MW-18Mar dataset.}
	\label{table:comp_weight_a}
 \centering
	\begin{tabular}[t]{@{}lcccccc@{}}
		\toprule
 & $\mathrm{AP}$ & $\mathrm{AP}_{50}$ & $\mathrm{AP}_{75}$ & $\mathrm{AP}_{\mathrm{S}}$ & $\mathrm{AP}_{\mathrm{M}}$ & $\mathrm{AP}_{\mathrm{L}}$ \\
 \midrule
 $\lambda_a = 1$ & 49.2 & 92.0 & 46.7 & 24.4 & 52.7 & 57.0 \\
 $\lambda_a = 0.1$ & 50.6 & 94.8 & 46.8 & 30.1 & 53.5 & 56.9 \\
		\bottomrule
	\end{tabular}
\end{table}
In our implementation, the angle loss is added to the original Deformable DETR. To analyze the effect of the weight for the angle loss, we compare the performance with $\lambda_a = 0.1$ and $\lambda_a = 1$. In this experiment, the performance of the detector trained with segmentation-based bounding boxes and pseudo-fisheye distortion is evaluated. Table~\ref{table:comp_weight_a} shows the comparison results. As shown in the table, AP, $\mathrm{AP}_{50}$, and $\mathrm{AP}_{75}$ with $\lambda_a = 0.1$ are better than those with $\lambda_a = 1$. We further compare the AP of the bounding boxes grouped by their areas. $\mathrm{AP}_{\mathrm{S}}$, $\mathrm{AP}_{\mathrm{M}}$, and $\mathrm{AP}_{\mathrm{L}}$ are calculated with bounding boxes whose areas are smaller than $64^2$, areas are between $64^2$ and $96^2$, and areas are larger than $96^2$, respectively. $\mathrm{AP}_{\mathrm{S}}$ and $\mathrm{AP}_{\mathrm{M}}$ with $\lambda_a = 0.1$ are better than those with $\lambda_a = 1$, while $\mathrm{AP}_{\mathrm{L}}$ is almost same for both values of $\lambda_a$. This is probably because the angle loss with the large weight overwhelms the $L_1$ box loss for small boxes. The $L_1$ box losses of small boxes tend to be smaller than those of large ones, and as a result, have a small impact on both the Hungarian matching and training losses with the large weight value for the angle loss, causing inappropriate matching and learning. With the small weight value for the angle loss, all the losses are balanced relatively well, resulting in better performance.

\subsubsection{Analysis of position}
\begin{figure}[tb]
	\centering
	\begin{minipage}[b]{1.0\linewidth}
	\centering
	\includegraphics[keepaspectratio,width=\linewidth]{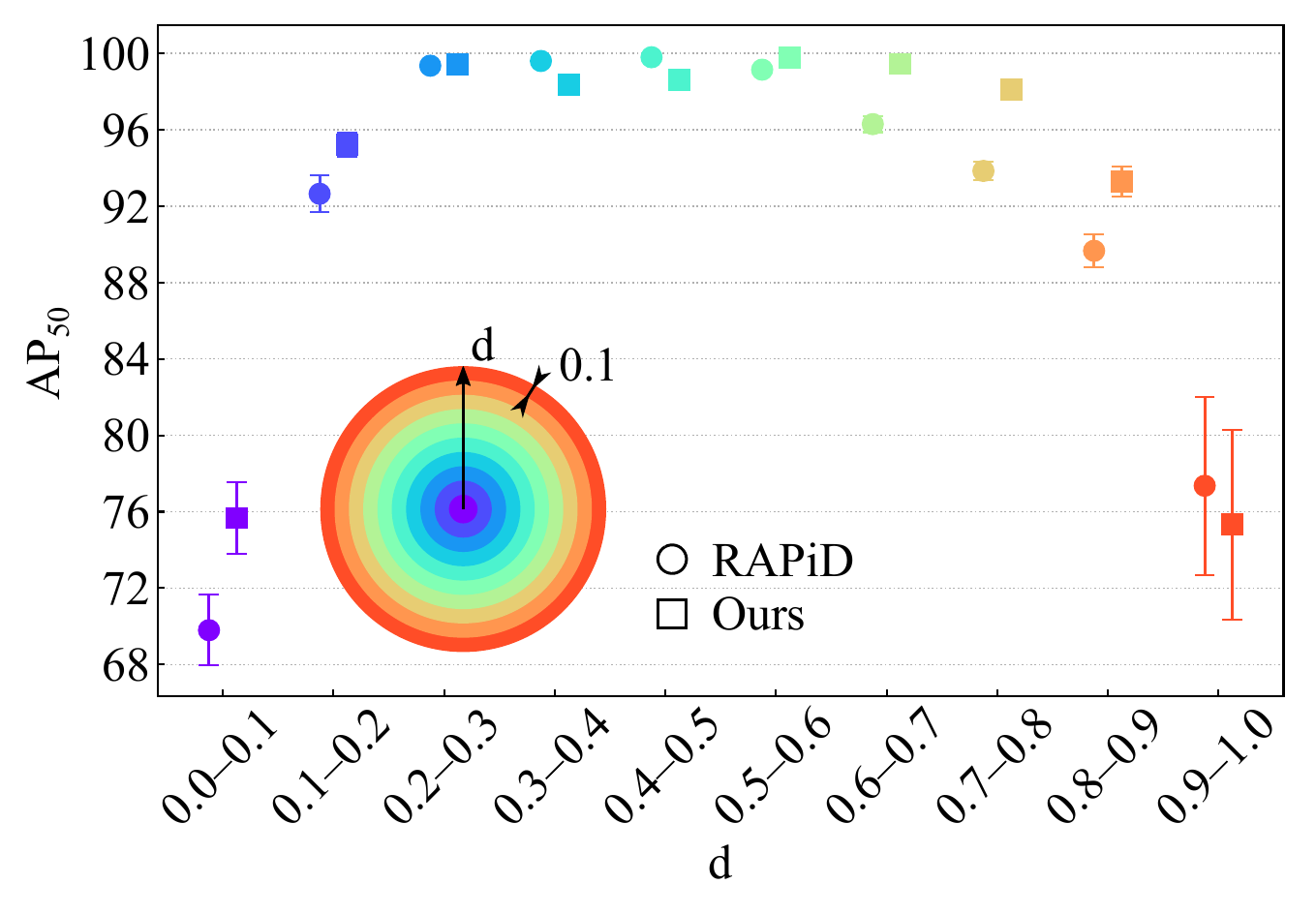}
	\subcaption{$\mathrm{AP}_{50}$ on each distance.}\label{fig:ap_r}
	\end{minipage}\\
	\begin{minipage}[b]{1.0\linewidth}
	\centering
	\includegraphics[keepaspectratio,width=\linewidth]{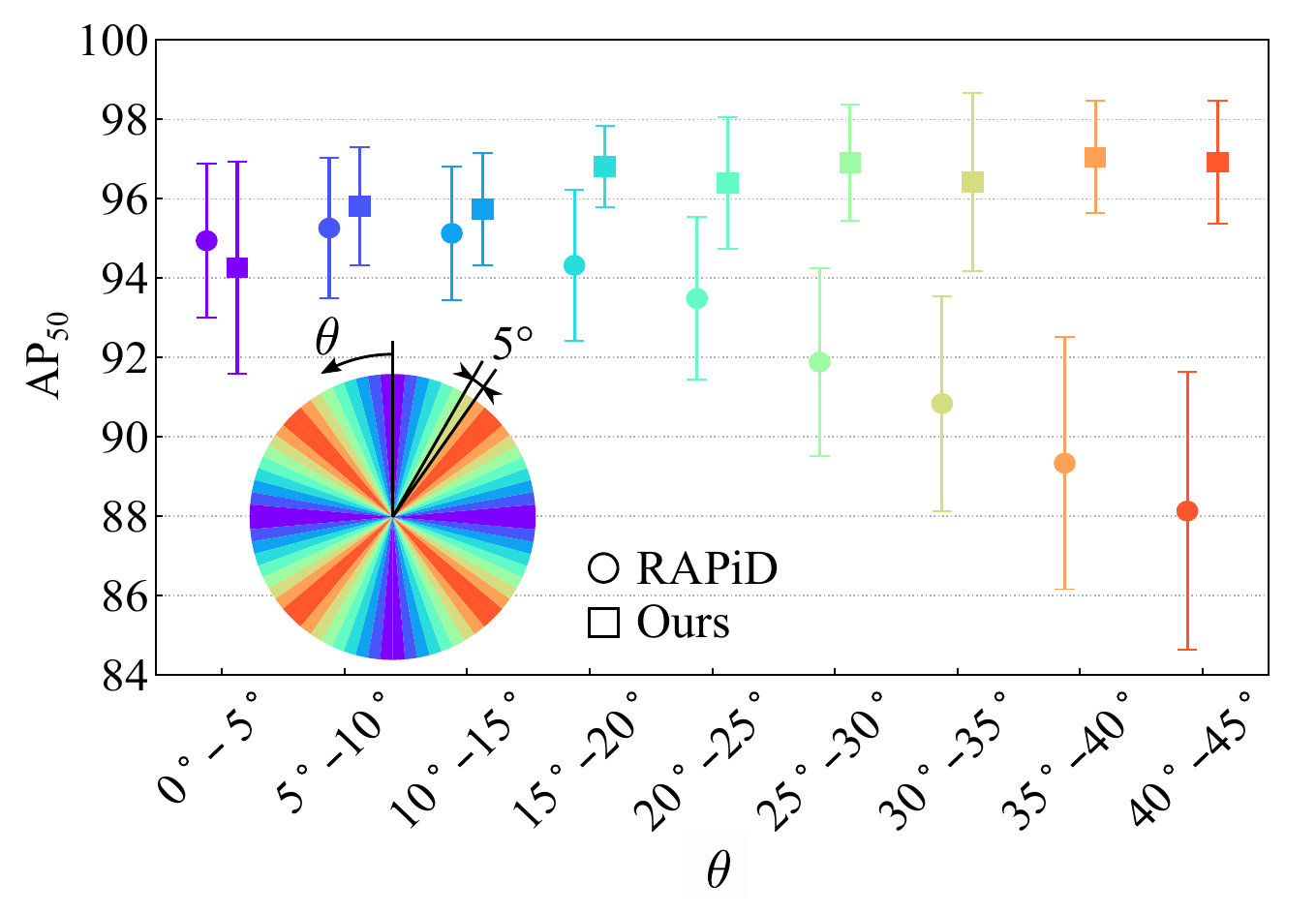}
	\subcaption{$\mathrm{AP}_{50}$ on each angle.}\label{fig:ap_theta}
	\end{minipage}
	\caption{$\mathrm{AP}_{50}$ on each position for the MW-18Mar dataset. The bars in each graph express standard errors. The marker colors denote the positions in each circle. (RAPiD: Trained with COCO's original bounding boxes, Ours: Trained with bounding boxes generated from segmentation annotations and added pseudo-fisheye distortion.)}\label{fig:ap}
\end{figure}   

To further analyze the detection results of the baseline and proposed methods, we conduct an analysis of the relationship between $\mathrm{AP}_{50}$ and the positions of the pedestrians in the images of the MW-18Mar dataset. For the baseline method, RAPiD is used, which shows the best results in the baseline methods, and for our method, the detector trained with segmentation-based bounding boxes and pseudo-fisheye distortion is used. We remove the dataset bias on the number of pedestrians and detection difficulty for each angle by rotating the images at 5-degree intervals and calculating $\mathrm{AP}_{50}$ for each interval. The means and standard errors of $\mathrm{AP}_{50}$ for all the rotation intervals are illustrated in Fig.~\ref{fig:ap}.

Figure~\ref{fig:ap_r} shows the relationship between $\mathrm{AP}_{50}$ and distances from image centers. As shown in the figure, both RAPiD and the proposed method demonstrate relatively low $\mathrm{AP}_{50}$ at the center and edge of the field of vision. The reason for the low $\mathrm{AP}_{50}$ at the center is that only heads and parts of bodies are taken there. Because that kind of appearance is rare in the COCO dataset, the detectors cannot detect pedestrians at the center. The reason for the low $\mathrm{AP}_{50}$ at the edge is that the detectors cannot detect pedestrians with tiny scales. This is the limitation of current detectors. In overall distances, the proposed method demonstrates almost the same or better performance than RAPiD, indicating the effectiveness of the proposed method.

Figure~\ref{fig:ap_theta} shows the relationship between $\mathrm{AP}_{50}$ and angles. As illustrated in the figure, the performance of RAPiD is degraded at angles of more than 15 degrees, while that of the proposed method is almost the same at any angle. In large angles, the appearance of pedestrians is typically inclined, which makes RAPiD difficult to tightly fit bounding boxes as illustrated in Fig.~\ref{fig:quality}. This is the reason for the degradation. The proposed detector differs in that it can tightly fit boxes to pedestrians at any angle, and hence demonstrates stable performance at any angle.

\subsection{Fine-tuning}

\begin{figure}
	\centering
	\begin{minipage}[b]{1.0\linewidth}
	\centering
	\includegraphics[keepaspectratio,width=\linewidth]{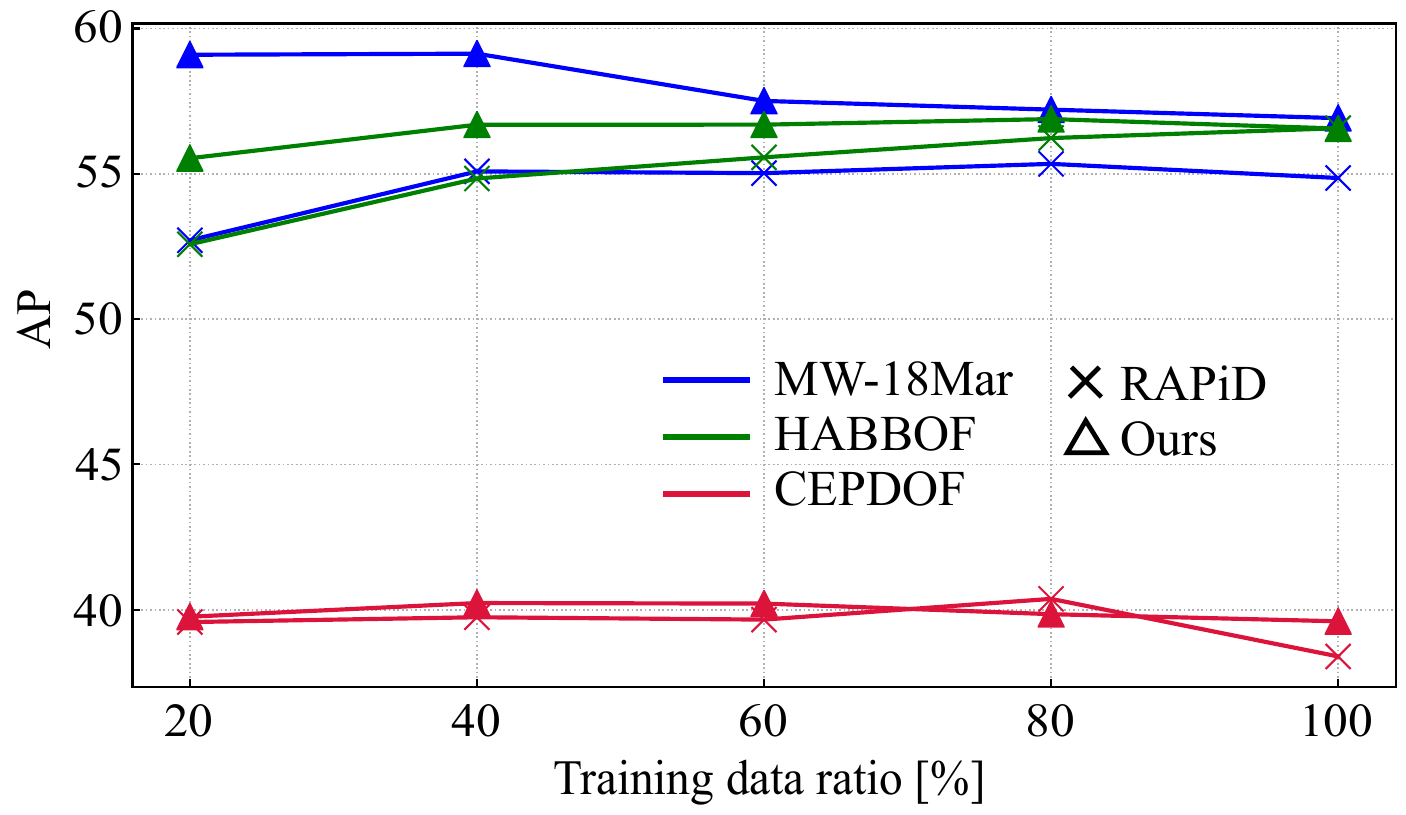}
	\subcaption{$\mathrm{AP}$ on various training data ratios.}\label{fig:finetune_ap_mod}
	\end{minipage}
	\begin{minipage}[b]{1.0\linewidth}
	\centering
	\includegraphics[keepaspectratio,width=\linewidth]{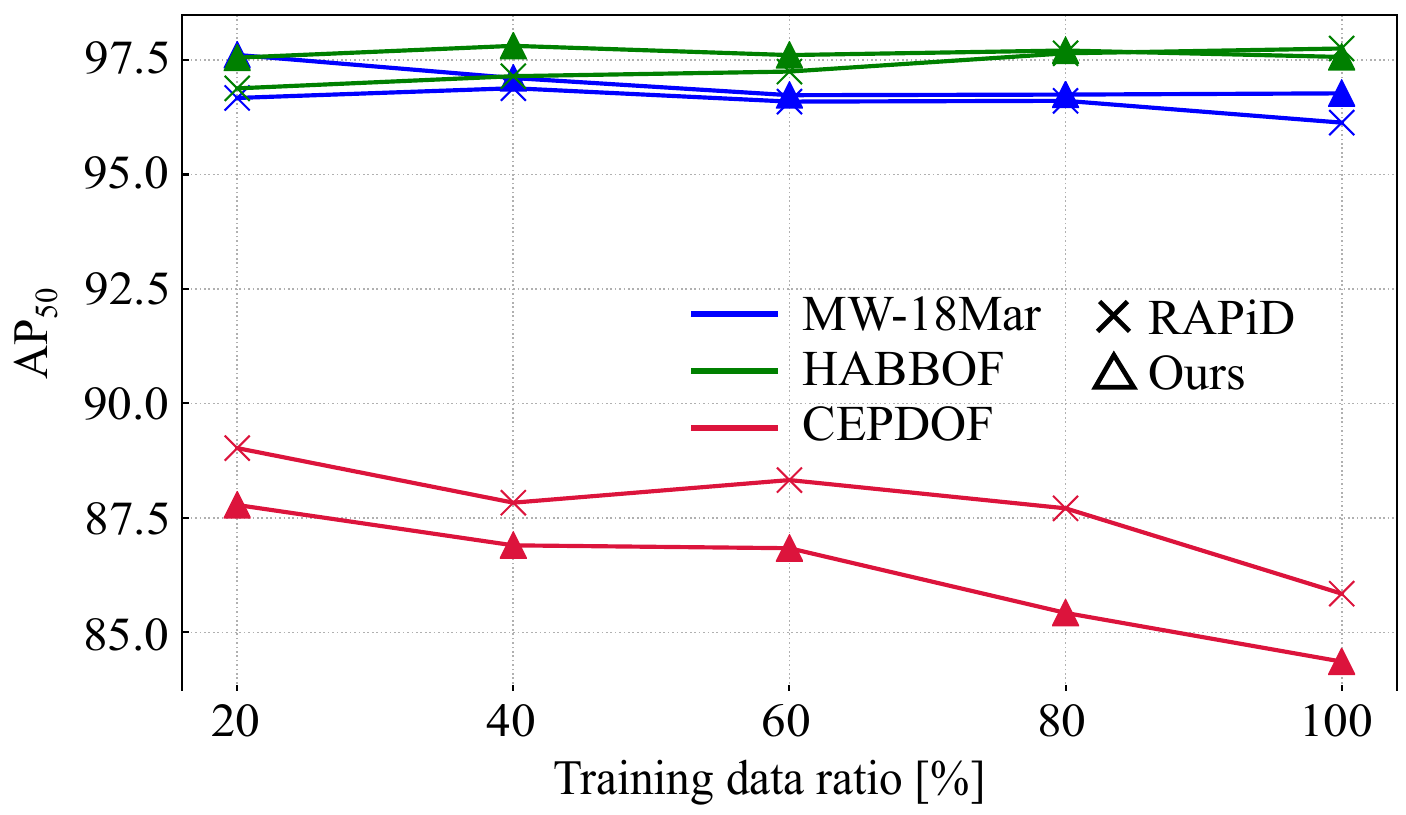}
	\subcaption{$\mathrm{AP}_{50}$ on various training data ratios.}\label{fig:finetune_ap50_mod}
	\end{minipage}\\
	\begin{minipage}[b]{1.0\linewidth}
	\centering
	\includegraphics[keepaspectratio,width=\linewidth]{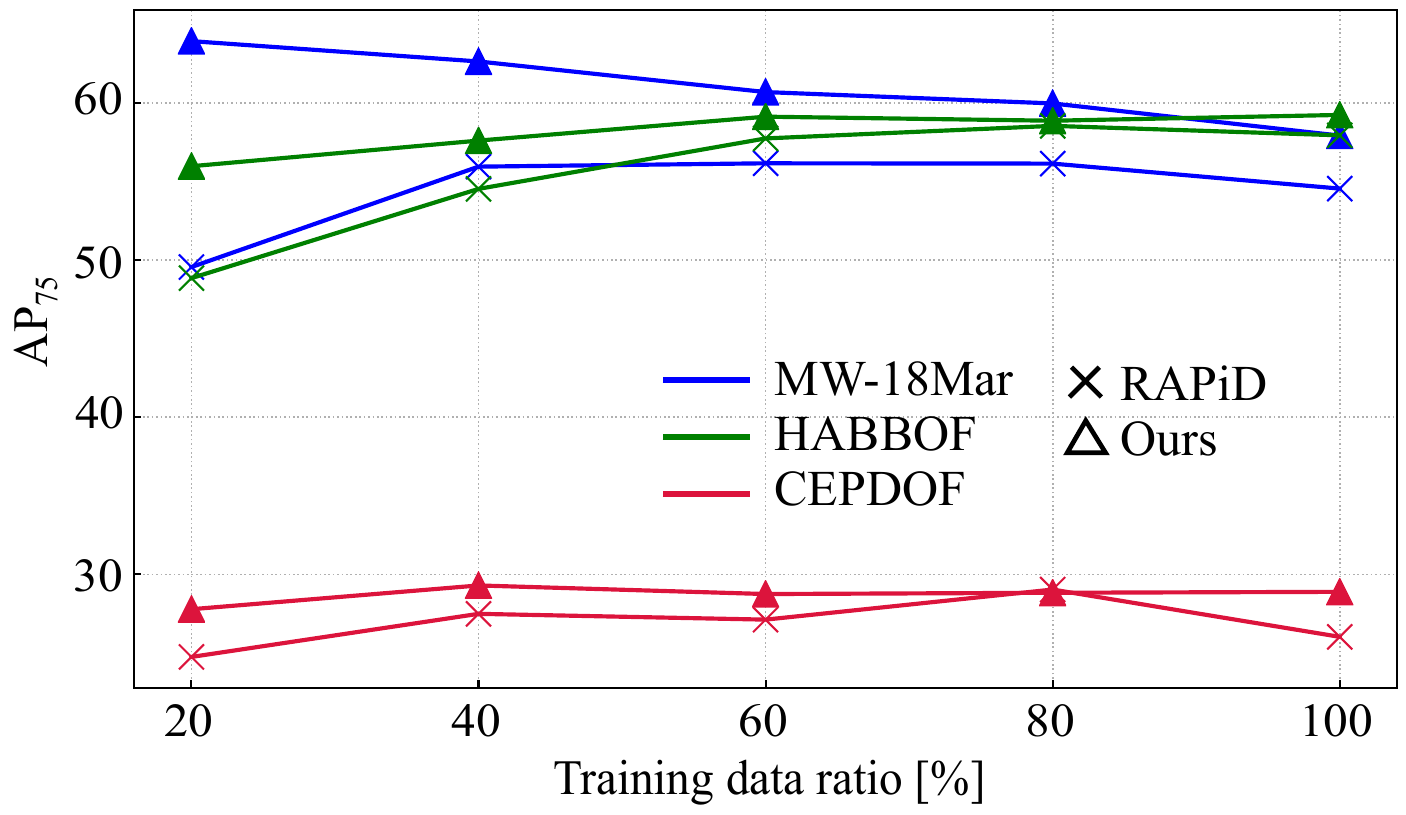}
	\subcaption{$\mathrm{AP}_{75}$ on various training data ratios.}\label{fig:finetune_ap75_mod}
	\end{minipage}
	\caption{APs on various training data ratios. Blue, green, and red colors illustrate the results of the MW-18Mar, HABBOF, and CEPDOF datasets, respectively. The cross and triangle markers show the RAPiD's results and ours, respectively. The pre-trained model of ours typically shows better performance than that of RAPiD with a small amount of omnidirectional training data, which indicates that our method is effective for generating pre-trained models for omnidirectional pedestrian detection.}\label{fig:finetune}
\end{figure}

To compare the performance as a pre-trained model for omnidirectional pedestrian detection, we fine-tune the detectors of RAPiD and the proposed method with omnidirectional images and evaluate each performance. We analyze the effect of the amount of omnidirectional training data by changing the amount from \SI{20}{\percent} to \SI{100}{\percent} of the entire images in the training datasets and evaluating the performance. Since our objective is to reduce the annotation costs for omnidirectional pedestrian detection, it is better to show high performance with a small amount of omnidirectional training data. Following Duan \textit{et al.} \cite{duan_cvprw2020}, we cross-validate the detectors on the three omnidirectional pedestrian detection datasets, i.e., two are used for training and the remaining one for testing. Two low-light scenes of the CEPDOF dataset are excluded during training.

Figure~\ref{fig:finetune} shows the evaluation results. The detector pre-trained with the proposed method typically demonstrates better performance especially in $\mathrm{AP}_{75}$ than that pre-trained with RAPiD when a small amount of omnidirectional training data is provided. These results indicate that our method has an advantage over RAPiD for pre-training omnidirectional pedestrian detectors that can precisely fit bounding boxes to pedestrians. 

The performance of the detector pre-trained with the proposed method is improved from the performance in Table~\ref{table:comp} by fine-tuning with omnidirectional images. However, when the amount of the training data increases, the performance does not change or is even degraded. These results indicate that the detector is well pre-trained, and the over-fitting occurs when the amount of training data increases.

\section{Conclusion}
In this paper, we proposed a segmentation-based bounding box generation method to enhance the performance of omnidirectional pedestrian detectors without using omnidirectional images for training. This method enables the detectors to tightly fit boxes to pedestrians and hence outperforms a conventional method, especially for the AP of high IoU thresholds. We also proposed a pseudo-fisheye distortion augmentation to further enhance the performance. This method enables perspective images to imitate omnidirectional images and thus improves the robustness to the distortion of omnidirectional images. Experimental results show that the proposed detector successfully fits boxes to pedestrians and achieves substantial performance improvement.

\section*{Declarations}
\subsection*{Funding}
No funding was received for conducting this study.

\subsection*{Competing interests}
The authors are employed by the company Hitachi, Ltd.

\subsection*{Availability of data and material}
The data that support the findings of this study are openly available on the dataset authors' websites listed below.
\begin{description}[font=\bfseries,style=unboxed,leftmargin=0cm]
	\item[COCO] \url{https://cocodataset.org/}.
	\item[MW-18Mar] \url{http://www2.icat.vt.edu/mirrorworlds/challenge/index.html}.
	\item[HABBOF] \url{http://vip.bu.edu/projects/vsns/cossy/datasets/habbof/}.
	\item[CEPDOF] \url{http://vip.bu.edu/projects/vsns/cossy/datasets/cepdof/}.
 \end{description}

\bibliography{main}

\end{document}